\documentclass[10pt,twocolumn,letterpaper]{article}

\usepackage{cvpr}
\usepackage{times}
\usepackage{epsfig}
\usepackage{graphicx}
\usepackage{amsmath}
\usepackage{amssymb}
\usepackage{mathtools}
\usepackage{algorithm}
\usepackage{algorithmic}
\usepackage{graphics}
\usepackage{epstopdf}
\usepackage{subfigure}
\usepackage{multirow}
\usepackage{booktabs}
\usepackage{url}
\usepackage{amsthm}

\usepackage{makecell}
\usepackage{ragged2e}

\usepackage[breaklinks=true,bookmarks=false]{hyperref}

\cvprfinalcopy 

\def\cvprPaperID{****} 

\begin{document}

\title{Towards Optimal Structured CNN Pruning via Generative Adversarial Learning}

\author{Shaohui Lin$^{1}$, Rongrong Ji$^{1}$\thanks{Corresponding author.}, Chenqian Yan$^{1}$, Baochang Zhang$^{2}$, \\ Liujuan Cao$^{1}$, Qixiang Ye$^{3}$, Feiyue Huang$^{4}$, David Doermann$^{5}$\\
$^1$Fujian Key Laboratory of Sensing and Computing for Smart City, School of Information Science \\and Engineering, Xiamen University, 361005, China\\
$^2$Beihang University, China,
$^3$University of Chinese Academy of Sciences, China\\
$^4$BestImage, Tencent Technology (Shanghai) Co.,Ltd, China,
$^5$University at Buffalo, USA\\
{\tt\small \{Shaohuilin007, cherrycherryan\}@gmail.com, \{rrji, caoliujuan\}@xmu.edu.cn} \\ {\tt\small bczhang@buaa.edu.cn, 
 qxye@ucas.ac.cn, garyhuang@tencent.com, doermann@buffalo.edu}}
\maketitle


\begin{abstract}
Structured pruning of filters or neurons has received increased focus for compressing convolutional neural networks.
Most existing methods rely on multi-stage optimizations in a layer-wise manner for iteratively pruning and retraining which may not be optimal and may be computation intensive.
Besides, these methods are designed for pruning a specific structure, such as filter or block structures  without jointly pruning heterogeneous structures. In this paper, we propose an effective structured pruning approach that jointly prunes filters as well as other structures in an end-to-end manner. To accomplish this, we first introduce a soft mask to scale the output of these structures by defining a new objective function with sparsity regularization to align the output of baseline and network with this mask. We then effectively solve the optimization problem by generative adversarial learning (GAL), which learns a sparse soft mask in a label-free and an end-to-end manner. By forcing more scaling factors in the soft mask to zero, the fast iterative shrinkage-thresholding algorithm (FISTA) can be leveraged to fast and reliably remove the corresponding structures. Extensive experiments demonstrate the effectiveness of GAL on different datasets, including MNIST, CIFAR-10 and ImageNet ILSVRC 2012. For example, on ImageNet ILSVRC 2012, the pruned ResNet-50 achieves 10.88\% Top-5 error and results in a factor of $3.7\times$ speedup. This significantly outperforms state-of-the-art methods.
\end{abstract}


\section{Introduction}
\label{intro}
Convolutional neural networks (CNNs) have achieved state-of-the-art accuracy in computer vision tasks such as image recognition \cite{krizhevsky2012imagenet,simonyan2015very,szegedy2015going,he2016deep,huang2017densely} and object detection \cite{girshick2014rich,girshick2015fast,ren2015faster}. However, the success of CNNs is often accompanied by significant computation and memory consumption that restricts their usage on resource-limited devices, such as mobile or embedded devices. To address these issues, techniques have been proposed for CNN compression such as low-rank decomposition \cite{denton2014exploiting,zhang2015efficient,lin2016towards,lin2018holistic}, parameter quantization \cite{rastegari2016xnor,jacob2018quantization,zhuang2018towards}, knowledge distillation \cite{hinton2015distilling,romero2015fitnets} and network pruning \cite{han2016deep,li2017pruning,lin2018accelerating,huang2018data,he2018pruning,lin2019towards}. Network pruning has received a great deal of research focus demonstrating significant compression and acceleration of CNNs in practice.

\begin{figure*}[t]

\begin{center}
\includegraphics[scale=0.4]{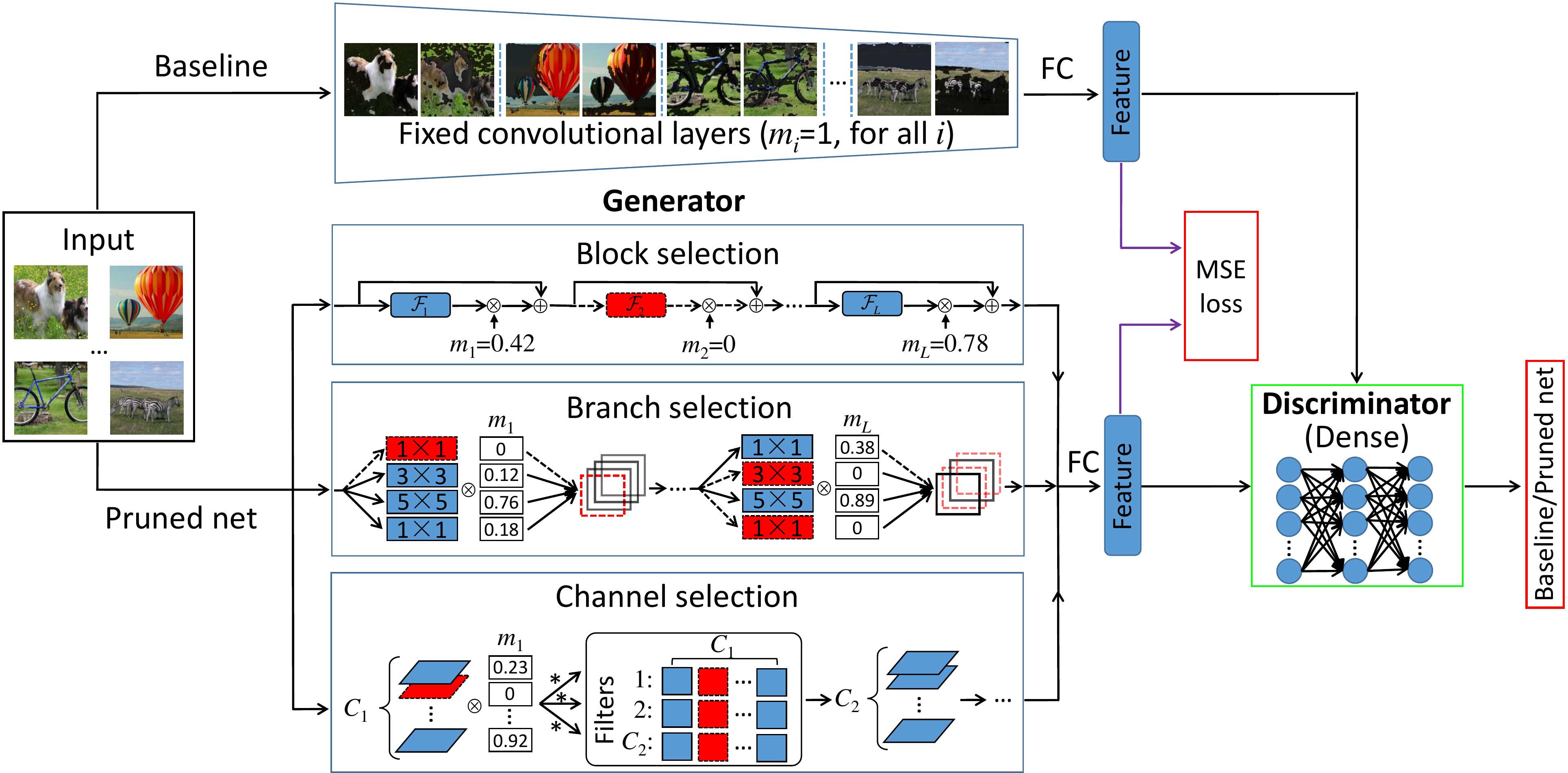}
\end{center}
\vspace{-0.5em}
\caption{An illustration of GAL. Blue solid block, branch and channel elements are active, while red dotted elements are inactive and can be pruned since their corresponding scaling factors in the soft mask are 0. (This figure is best viewed in color and zoomed in.)}
\label{fig1}
\end{figure*}

Network pruning can be categorized into either non-structured or structured. Non-structured pruning or fine-grained pruning \cite{han2015learning,han2016deep,lecun1989optimal,hassibi1993second}, directly pruning weights independently in each layer to achieve higher sparsity for the remaining parameters. However, it generally causes irregular memory access that adversely impacts the efficiency of online inference. Under such a circumstance, specialized hardware \cite{han2016eie} or software \cite{park2017faster} accelerators are required to further speedup the sparse CNNs. 
Structured or coarse-grained pruning \cite{li2017pruning,wen2016learning,huang2018data,luo2017ThiNet,he2017channel,liu2017learning} aims to remove structured weights, including 2D kernels, filters or layers, and does not require specialized hardware/software packages to be efficiently implemented. However, there exists several open issues in the existing structured pruning. 
(1) \emph{Efficiency}: The existing approaches typically adopt iterative pruning and retraining with multi-stage optimizations in a layer-wise manner. For instance, Luo \emph{et al.} \cite{luo2017ThiNet} and He \emph{et al.} \cite{he2017channel} proposed to prune filters and the corresponding feature maps by considering statistics computed from the next layer in a greedy layer-wise manner. Magnitude-based pruning methods employ the $\ell_1$-norm of filter \cite{li2017pruning} or the sparsity of feature map \cite{hu2016network} to determine the importance of the filter. They then iteratively prune the ``least important'' filters and retrain the pruned network layer-by-layer. 
(2) \emph{Slackness}: Existing approaches lack slackness in hard filter pruning. For instance, Lin \emph{et al.} \cite{lin2018accelerating} learned a global mask with binary values to determine the saliency of filters, and pruned the redundant filters by masking out the corresponding mask as 0. 
However, such a \emph{hard} filter pruning using binary masks results in the difficulty to solve the optimization problem.
(3) \emph{Label dependency}: Most existing pruning approaches rely on a pre-defined set of labels to learn the pruning strategy. For instance, group sparsity with $\ell_{2,1}$-regularization on the filters \cite{wen2016learning} and sparsity with $\ell_1$-regularization on the scaling parameters \cite{huang2018data,liu2017learning,ye2018rethinking} were utilized to generate a sparse network by training with class labels. These training schemes cannot be directly used in scenarios without labels.

To address these issues, we propose an effective structured pruning approach to prune heterogeneous redundant structures (including channels/filters, branches and blocks) in an end-to-end manner without iteratively pruning and retraining. 
Unlike previous approaches of hard and label-dependent pruning, we propose a label-free generative adversarial learning (GAL) to prune the network with a sparse soft mask, which scales the output of specific structures to be zero. Fig. \ref{fig1} depicts the workflow of the proposed approach. 
We first initialize a pruned network with the same weights as the baseline or the pre-trained network, and initialize a soft mask randomly after each structure.
We then construct a new objective function with $\ell_1$-regularization on the soft mask to align the outputs of the baseline and the pruned network. 
To effectively solve the optimization problem, the proposed label-free generative adversarial learning learns the pruned network with this sparse soft mask in an end-to-end manner inspired by Generative Adversarial Networks (GANs) \cite{goodfellow2014generative}. 
The optimization is playing a two-player game where the generator is the pruned network, and the discriminator distinguishes whether the input is from the output of the baseline or from the pruned network. This forces the two outputs to be close to each other. We introduce an adversarial regularization on the discriminator to help the pruned network to compete with the discriminator. By forcing more scaling factors in the soft mask to zero, we can leverage the fast iterative shrinkage-thresholding algorithm \cite{beck2009fast,goldstein2014field} to reliably remove the corresponding structures.   

Our main contributions are summarized as follows:
\begin{itemize}
\vspace{-.5em}
\item[1.] We propose a generative adversarial learning (GAL) to effectively conduct structured pruning of CNNs. It is able to jointly prune redundant structures, including filters, branches and blocks to improve the compression and speedup rates.
\vspace{-.5em}
\item[2.] Adversarial regularization is introduced to prevent a trivially-strong discriminator, soft mask is used to solve the slackness of hard filter pruning, and FISTA is employed to fast and reliably remove the redundant structures.
\vspace{-.5em}
\item[3.] Extensive experiments demonstrate the superior performance of our approach. On ImageNet ILSVRC 2012 \cite{russakovsky2015imagenet}, the pruned ResNet-50 achieves 10.88\% Top-5 error with a factor of $3.7\times$ speedup outperforming state-of-the-art methods.   
\end{itemize}

\section{Related Work}
\textbf{Network Pruning:} Network pruning focuses on removing network connections in non-structured or structured manner as introduced in Section \ref{intro}. Early work in non-structured pruning \cite{lecun1989optimal} and \cite{hassibi1993second} proposed a saliency measurement to remove redundant weights determined by the second-order derivative matrix of the loss function \emph{w.r.t.} the weights. Han \emph{et al.} \cite{han2015learning,han2016deep} proposed an iterative thresholding to remove unimportant weights with small absolute values. Guo \emph{et al.} \cite{guo2016dynamic} proposed a connection splicing to avoid incorrect weight pruning, which can reduce the accuracy loss of the pruned network. 
In contrast, structured pruning can reduce the network size and achieve fast inference without specialized packages. Li \emph{et al.} \cite{li2017pruning} proposed a magnitude-based pruning to remove filters and their corresponding feature maps by calculating the $\ell_1$-norm of filters in a layer-wise manner. A Taylor expansion based criterion was proposed in \cite{molchanov2017pruning} to iteratively prune one filter and then fine-tune the pruned network. This is, however, prohibitively costly for deep networks. Unlike these multi-stage and layer-wise pruning methods, our method prunes the network with the sparse soft mask by an end-to-end training that achieves much better results as quantitatively shown in our experiments. 

Recently, binary masks have been proposed to guide filter pruning. Yu \emph{et al.} \cite{yu2018nisp} proposed a Neuron Importance Score Propagation (NISP) to optimize the reconstruction error of the ``final response layer'' and propagate an ``importance score'' to each node, \emph{i.e.}, 1 for important nodes, and 0 otherwise. 
Lin \emph{et al.} \cite{lin2018accelerating} directly learned a global mask with binary values, and pruned the filters whose mask values are 0. However, such a hard filter pruning lacks effectiveness and slackness, due to the NP-hard optimization caused by using the binary mask. Our method slacks the binary mask to the soft one, which largely improves the flexibility and accuracy. 

In line with our work, sparse scaling parameters \cite{liu2017learning,ye2018rethinking} in batch normalization (BN) or in the specific structures \cite{huang2018data} were obtained by supervised training with a class-labelled dataset. In contrast, our approach obtains the sparse soft mask with label-free data and can transfer to other scenarios with unseen labels.

\textbf{Neural Architecture Search: } While state-of-the-art CNNs with compact architectures have been explored with hand-crafted design \cite{howard2017mobilenets,zhang2018shufflenet,zhang2017interleaved}, automatic search of neural architectures is also becoming popular. Recent work on searching models with reinforcement learning \cite{baker2017designing,zoph2017neural,zoph2018learning,he2018amc} or genetic algorithms \cite{real2017large,xie2017genetic} greatly improve the performance of neural networks. However, the search space of these methods is extremely large, which requires significant computational overhead to search and select the best model from hundreds of models. 
In contrast, our method learns a compact neural architecture by a single training, which is more efficient. Group sparsity regularization on filters \cite{lebedev2016fast} or multiple structures including filter shapes and layers \cite{wen2016learning} has been proposed to sparsify them during training. This is also less efficient and cannot reliably remove the sparse structures since only stochastic gradient descent is used.

\textbf{Knowledge Distillation:} The proposed generative adversarial learning for structured pruning is also related to knowledge distillation (KD) to a certain extent. KD transfers knowledge from the teacher to the student using different kinds of knowledge (\emph{e.g.}, dark knowledge \cite{hinton2015distilling,romero2015fitnets} and attention \cite{zagoruyko2017paying}). Hinton \emph{et al.} \cite{hinton2015distilling} introduced dark knowledge for model compression, which uses the softened final output of a complicated teacher network to teach a small student network. Romero \emph{et al.} \cite{romero2015fitnets} proposed FitNets to train the student network by combining dark knowledge and the knowledge from the teacher's hint layer. Zagoruyko \emph{et al.} \cite{zagoruyko2017paying} transferred the knowledge from attention maps from the teacher's hidden layer to improve the performance of a student network. 
Unlike other methods, we do not require labels to train the pruned network. Furthermore, we directly copy the architecture of the student network from the teacher without being designed by experts, and then automatically learn how to prune the student network. 

Note that our approach is orthogonal to other compression approaches, such as low-rank decomposition \cite{denton2014exploiting,lebedev2015speeding,kim2016compression,zhang2015efficient,lin2018holistic}, or parameter quantization \cite{rastegari2016xnor,jacob2018quantization,zhuang2018towards}.
We can integrate our approach into the above methods to achieve higher compression and speedup rates.

\section{Our Method}

\subsection{Notations and Preliminaries}
As illustrated in Fig. \ref{fig1}, we define an original pre-trained network as the baseline $f_b(x,\mathcal{W}_B)$ and the network with soft mask as the pruned network $f_g(x,\mathcal{W}_{G}, \textbf{m})$, where $x,\mathcal{W}_B$ and $\mathcal{W}_G$ are an input image, all weights in the baseline and all weights in the pruned network, respectively. 
$\mathcal{W}_G^l$ represents the convolutional filters or neurons at the $l$-th layer in $\mathcal{W}_G$ with a tensor size of $H_l\times W_l\times C_l\times N_l$. 
$\textbf{m}\in\mathbb{R}^s$ is the soft mask after each structure, where $s$ is the number of structures we consider to prune, and $m_i$ refers to the $i$-th element of $\textbf{m}$. 
Since the baseline is fixed and not updated during training, we select its final output (before the probabilistic ``softmax'') as the supervised feature $f_b(x)$ to train the pruned network.
We also extract the feature from the final output of the pruned network, which is denoted as $f_g(x)$. Different from $f_b(x)$, $f_g(x)$ requires updating with soft mask $\textbf{m}$ and weights $\mathcal{W}_{G}$ to approximate $f_b(x)$. 

\subsection{Formulation}
We aim to learn a soft mask to remove the corresponding structures including channels, branches and blocks, while regaining close to the baseline accuracy. Inspired by knowledge distillation \cite{hinton2015distilling}, we train the pruned network with $\ell_1$-regularization on the soft mask to mimic the baseline by aligning their outputs. We obtain the pruned network by generative adversarial learning. The discriminator $D$ with weights $\mathcal{W}_D$ is introduced to distinguish the output of baseline or pruned network, and then the generator (\emph{i.e.}, the pruned network) $G$ with weights $\mathcal{W}_G$ and soft mask $\mathbf{m}$ is learned together with $D$ by using the knowledge from supervised features of baseline.
Therefore, $\mathcal{W}_G, \mathbf{m}$ and $\mathcal{W}_D$ are learned by solving the optimization problem as follows:
\vspace{-.5em}
\begin{small}
\begin{equation}
\label{eq1}
\begin{split}
\arg\min_{\mathcal{W}_G,\mathbf{m}}\max_{\mathcal{W}_D} \, & \mathcal{L}_{Adv}(\mathcal{W}_G,\mathbf{m},\mathcal{W}_D)+\mathcal{L}_{data}(\mathcal{W}_G,\mathbf{m}) \\
& +\mathcal{L}_{reg}(\mathcal{W}_G,\mathbf{m},\mathcal{W}_D),
\end{split}
\end{equation}
\end{small}

\noindent where $\mathcal{L}_{Adv}(\mathcal{W}_G,\mathbf{m},\mathcal{W}_D)$ is the adversarial loss to train the two-player game between the baseline and the pruned network that compete with each other. This is defined as:
\begin{footnotesize}
\begin{equation}
\label{eq2}
\begin{split}
\mathcal{L}_{Adv}&(\mathcal{W}_G,\mathbf{m},\mathcal{W}_D) = \mathbb{E}_{f_b(x)\sim p_b(x)}\Big[log\big(D(f_b(x),\mathcal{W}_D)\big)\Big]\\
& +\mathbb{E}_{f_g(x,z)\sim (p_g(x), p_z(z))}\Big[log\big(1-D(f_g(x,z),\mathcal{W}_D)\big)\Big],\\
\end{split}
\end{equation}
\end{footnotesize}

\vspace{-.5em}
\noindent where $p_b(x)$ and $p_g(x)$ represent the feature distributions of the baseline and the pruned network, respectively. $p_z(x)$ corresponds to the prior distribution of noise input $z$. Inspired by \cite{isola2017image}, we use the dropout as the noise input $z$ in the pruned network. This dropout is active only while updating the pruned network. For notation simplicity, we omit $z$ in $f_g(x,z)$. 

In addition, $\mathcal{L}_{data}(\mathcal{W}_G,\mathbf{m})$ is the data loss between output features from both the baseline and the pruned network, which is used to align the outputs of these two networks. Therefore, the data loss can be expressed by MSE loss:
\vspace{-.5em}
\begin{small}
\begin{equation}
\label{eq3}
\mathcal{L}_{data}(\mathcal{W}_G,\mathbf{m})=\frac{1}{2n}\sum_x\big\|f_b(x)-f_g(x,\mathcal{W}_G,\mathbf{m})\big\|_2^2,
\end{equation}
\end{small}

\vspace{-1em}
\noindent where $n$ is the number of the mini-batch size. 

Finally, $\mathcal{L}_{reg}(\mathcal{W}_G,\mathbf{m},\mathcal{W}_D)$ is a regularizer on $\mathcal{W}_G,\mathbf{m}$ and $\mathcal{W}_D$, which can be split into three parts as follows:
\begin{small}
\begin{equation}
\label{eq4}
\mathcal{L}_{reg}(\mathcal{W}_G,\mathbf{m},\mathcal{W}_D)=\mathcal{R}(\mathcal{W}_G)+\mathcal{R}_{\lambda}(\mathbf{m})+\mathcal{R}(\mathcal{W}_D),
\end{equation}
\end{small}

\vspace{-1.5em}
\noindent where $\mathcal{R}(\mathcal{W}_G)$ is the weight decay $\ell_2$-regularization in the pruned network, which is defined as $\frac{1}{2}\|\mathcal{W}_G\|_2^2$. 
$\mathcal{R}_{\lambda}(\mathbf{m})$ is a sparsity regularizer for $\mathbf{m}$ with parameter $\lambda$. If $m_i=0$, we can reliably remove the corresponding structure as its corresponding output has no contribution to the subsequent computation. In practice, we employ the widely-used $\ell_1$-regularization to constrain $\mathbf{m}$, which is defined as $\lambda\|\mathbf{m}\|_1$. 
$\mathcal{R}(\mathcal{W}_D)$ is a discriminator regularizer used to prevent the discriminator from dominating the training, while retaining the network capacity. In this paper, we introduce three kinds of discriminator regularizations including $\ell_{1}$-regularization, $\ell_{2}$-regularization and adversarial regularization. 
We add a negative sign in both $\ell_{1}$-regularization and $\ell_{2}$-regularization. This is different from the definition above, since $\mathcal{W}_D$ is updated by the maximization of Eq. (\ref{eq1}).
The adversarial regularization (AR) is defined as:
\begin{small}
\begin{equation}
\label{eq7}
\mathcal{R}(\mathcal{W}_D) = \mathbb{E}_{f_g(x)\sim p_g(x)}\big[log(D(f_g(x),\mathcal{W}_D))\big].
\end{equation}
\end{small}

\vspace{-1.4em}
\noindent We found the discriminator $D$ is updated only with correct prediction by using Eq. (\ref{eq2}), which leads to a less valuable gradient updating that the pruned network receives. 
Therefore, adversarial regularization is introduced to also update the discriminator $D$ with the features of pruned network produced by the baseline, and to extend the time of the two-player game to achieve more valuable gradients.

\begin{algorithm}[t]
\label{alg1}
\small
\caption{FISTA in GAN to solve Eq. (\ref{eq1})}
\renewcommand{\algorithmicrequire}{\textbf{Input:}} 
\renewcommand{\algorithmicensure}{\textbf{Output:}}
\begin{algorithmic}[1]
\REQUIRE 
Training data $\mathbf{X}=\{x^1,\cdots,x^n\}$ with $n$ samples,
baseline model $\mathcal{W}_B=\{\mathcal{W}_B^1,\cdots,\mathcal{W}_B^L\}$, sparsity factor $\lambda$, number of steps $i$ and $j$ to apply to the discriminator $D$ and generator $G$, learning rate $\eta$, maximum iterations $T$.
\ENSURE 
The weights $\mathcal{W}_G=\{\mathcal{W}_G^1,\cdots,\mathcal{W}_G^L\}$ and their soft masks $\mathbf{m}$. \\
\STATE
Initialize $\mathcal{W}_G=\mathcal{W}_B$, $\mathbf{m}\sim\mathcal{N}(0,1)$, and $t=1$.
\REPEAT 
\STATE 
 \textbf{for} $i$ steps \textbf{do} \\
 (\textbf{Fix $G$ and update $D$})
 \begin{itemize}
 \item  Forward pass baseline to sample minibatch of $s$ examples $\big\{f_b(x^1),\cdots, f_b(x^s)\big\}$.
 \item  Forward pass generator to sample minibatch of $s$ examples $\big\{f_g(x^1),\cdots, f_g(x^s)\big\}$.
 \item Remove term $\mathcal{L}_{data}(\mathcal{W}_G,\mathbf{m}),\mathcal{R}(\mathcal{W}_G)$ and $\mathcal{R}_{\lambda}(\mathbf{m})$, and solve the following optimization to update $D$:
\vspace{-.5em}
\begin{tiny}
\begin{equation}
\label{eq8}
\begin{split}
\arg\max_{\mathcal{W}_D} \, & \mathbb{E}_{f_b(x)\sim p_b(x)}\big[log(D(f_b(x),\mathcal{W}_D))\big] +\mathcal{R}(\mathcal{W}_D) \\
& +\mathbb{E}_{f_g(x)\sim p_g(x)}\big[log(1-D(f_g(x),\mathcal{W}_D))\big].
\end{split}
\end{equation}
\end{tiny}
 \end{itemize}
\vspace{-.5em}
\textbf{end for}\\ 
\STATE
 \textbf{for} $j$ steps \textbf{do} \\
 (\textbf{Fix $D$ and update $G$})
 \begin{itemize}
 \item  Forward pass baseline to sample minibatch of $s$ examples $\big\{f_b(x^1),\cdots, f_b(x^s)\big\}$.
 \item  Forward pass generator to sample minibatch of $s$ examples $\big\{f_g(x^1,z),\cdots, f_g(x^s,z)\big\}$ with dropout as noise input.
 \item Remove term $\mathcal{R}(\mathcal{W}_D)$ and $\mathbb{E}_{f_b(x)\sim p_b(x)}\big[log(D(f_b(x),\mathcal{W}_D))\big]$, and solve the following optimization to update $G$ by FISTA:
\vspace{-.5em}
\begin{tiny}
\begin{equation}
\label{eq9}
\begin{split}
\arg\min_{\mathcal{W}_G,\mathbf{m}} & \mathbb{E}_{f_g(x,z)\sim (p_g(x), p_z(z))}\big[log(1-D(f_g(x,z),\mathcal{W}_D))\big] \\
& +\frac{1}{2n}\sum_x\big\|f_b(x)-f_g(x,\mathcal{W}_G,\mathbf{m})\big\|_2^2 \\
& +\frac{1}{2}\|\mathcal{W}_G\|_2^2+\lambda\|\mathbf{m}\|_1.
\end{split}
\end{equation}
\end{tiny}
 \end{itemize}
\vspace{-.5em} 
\textbf{end for}\\
\UNTIL{convergence or $t$ reaches the maximum iterations $T$.} 
\end{algorithmic}
\label{alg1}
\end{algorithm}

\subsection{Optimization}
Following \cite{goodfellow2014generative}, Stochastic Gradient Descent (SGD) can be directly introduced to alternately update the discriminator $D$ and generator $G$ to solve the optimization problem in Eq. (\ref{eq1}). However, SGD is less efficient in convergence, and by using SGD we have observed non-exact zero scaling factors in the soft mask $\mathbf{m}$. 
We therefore need a threshold to remove the corresponding structures, whose scaling factors are lower than the threshold. By doing so, the accuracy of the pruned network is significantly lower than the baseline. To solve this problem, we introduce FISTA \cite{beck2009fast,goldstein2014field} into the GAN to effectively solve the optimization problem of Eq. (\ref{eq1}) via two alternating steps. Algorithm \ref{alg1} presents the optimization process.

First, we use SGD to optimize the weights $\mathcal{W}_D$ of the discriminator $D$ by ascending its stochastic gradient to solve Eq. (\ref{eq8}). The entire procedure mainly relies on the standard forward-backward pass.
Second, for better illustration, we shorten the first two terms of Eq. (\ref{eq9}) as $\mathcal{H}(\mathcal{W}_G, \mathbf{m})$, and we have:
\vspace{-.5em}
\begin{small}
\begin{equation}
\label{eq10}
\arg\min_{\mathcal{W}_G,\mathbf{m}} \mathcal{H}(\mathcal{W}_G, \mathbf{m})+\frac{1}{2}\|\mathcal{W}_G\|_2^2+\lambda\|\mathbf{m}\|_1.
\end{equation}
\end{small}

\vspace{-1.2em}
\noindent We solve the optimization problem of Eq. (\ref{eq10}) by alternately updating $\mathcal{W}_G$ and $\mathbf{m}$. (1) Fixing $\mathbf{m}$, we use SGD with momentum to update $\mathcal{W}_G$ by descending its gradient. (2) Fixing $\mathcal{W}_G$, the optimization of $\mathbf{m}$ is reformulated as:
\vspace{-.5em}
\begin{small}
\begin{equation}
\label{eq11}
\arg\min_{\mathbf{m}} \mathcal{H}(\cdot, \mathbf{m})+\lambda\|\mathbf{m}\|_1.
\end{equation}
\end{small}

\vspace{-1.2em}
\noindent Then $\mathbf{m}$ is updated by FISTA with the initialization of $\alpha_{(1)} = 1$:
\begin{small}
\begin{equation}
\label{eq12}
\alpha_{(k+1)}=\frac{1}{2}\Big(1+\sqrt{1+4\alpha_{(k)}^{2}}\Big),
\end{equation}
\end{small}
\begin{small}
\begin{equation}
\label{eq13}
\mathbf{y}_{(k+1)}=\mathbf{m}_{(k)}+\frac{\alpha_{(k)}-1}{\alpha_{(k+1)}}\big(\mathbf{m}_{(k)}-\mathbf{m}_{(k-1)}\big),
\end{equation}
\end{small}
\begin{scriptsize}
\begin{equation}
\label{eq14}
\mathbf{m}_{(k+1)}=\textbf{prox}_{\eta_{(k+1)}\lambda\|\cdot\|_1}\bigg(\mathbf{y}_{(k+1)}-\eta_{(k+1)}\frac{\partial\mathcal{H}(\cdot,\mathbf{y}_{(k+1)})}{\partial\mathbf{y}_{(k+1)}}\bigg),
\end{equation}
\end{scriptsize}

\vspace{-.8em}
\noindent where $\eta_{(k+1)}$ is the learning rate at the iteration $k+1$ and $\textbf{prox}_{\eta_{(k+1)}\lambda\|\cdot\|_1}(\mathbf{z}_i)=\text{sign}(\mathbf{z}_i)\circ(|\mathbf{z}_i|-\eta_{(k+1)}\lambda)_{+}$.

We solve these two steps by following stochastic methods with the mini-batches and set the learning rate $\eta$ with fixed-step updating. Moreover, we update $\small{\mathcal{W}_G,\mathbf{m}}$ and $\small{\mathcal{W}_D}$ at each iteration ($i=j=1$ in Algorithm \ref{alg1}).

\subsection{Structure Selection}
\label{SS}
To achieve flexible structure selection, we add a soft mask after the three different kinds of structures from coarse to fine-grained, including blocks, branches and channels, to remove the redundancy of different networks ResNets \cite{he2016deep}, GoogLeNet \cite{szegedy2015going} and DenseNets \cite{huang2017densely} as shown in Fig. \ref{fig1}. Furthermore, these structures can be integrated into each other for jointly learning. 

\textbf{Block Selection:}
For ResNets, the residual block contains the residual mapping with a large number of parameters and the shortcut connections with few parameters. This achieves high performance by skipping the computation of specific layers to overcome the degradation problem. 
The block is removed by setting the residual mapping to zero, but cannot cut off the information flow in ResNets. Therefore, block selection is significantly effective when applied in ResNets. The new residual block by adding the soft mask is formulated as:
\vspace{-.5em}
\begin{small}
\begin{equation}
\label{eq15}
\mathbf{z}^{i+1}=m_i\mathcal{F}\big(\mathbf{z}^i,\{\mathcal{W}_G^i\}\big)+\mathbf{z}^i,
\end{equation}
\end{small}

\vspace{-1.5em}
\noindent where $\mathbf{z}^i$ and $\mathbf{z}^{i+1}$ are the input and output of the $i$-th block, respectively. $\mathcal{F}$ is a residual mapping and $\{\mathcal{W}_G^i\}$ are weights of the $i$-th block. After optimization, we obtain a sparse soft mask $\mathbf{m}$, in which the $i$-th residual block can be pruned if $m_i=0$.

\textbf{Branch Selection.} 
Multi-branch networks such as GoogLeNet and ResNeXts \cite{xie2017aggregated} have been proposed to enhance the information flow to achieve high performance. Similar to ResNets, there is redundancy in the branch that can be removed entirely by setting the corresponding soft mask to 0. Likewise, this does not cut off the information flow in multi-branch networks. Taking GoogLeNet for instance, we can formulate the new inception module by adding the soft mask as follows:
\vspace{-.5em}
\begin{small}
\begin{equation}
\label{eq16}
\mathcal{T}(\mathbf{z})=\big[m_1\tau^1(\mathbf{z},\{\mathcal{W}_G^1\}),\cdots,m_c\tau^c(\mathbf{z},\{\mathcal{W}_G^c\})\big],
\end{equation}
\end{small}

\vspace{-1.5em}
\noindent where $[\cdot]$ represents concatenation operator. $\tau^i(\mathbf{z},\{\mathcal{W}_G^i\})$ is a transformation with all weights $\{\mathcal{W}_G^i\}$ at the $i$-th branch and $c$ is the number of branch in one inception module. We can reliably remove the $i$-th branch, which satisfies $m_i=0$ after optimization.

\textbf{Channel Selection:}
The channel is a basic element in all CNNs and has large amounts of redundancy. In our framework, we add the soft mask after input at the current layer (the output feature maps at the upper layer) to guide the input channel pruning at the current layer and the output channel pruning at the upper layer. Therefore, the formulation at the $l$-th layer is as follows:
\vspace{-.5em}
\begin{small}
\begin{equation}
\label{eq17}
\mathbf{z}_j^{l+1}=f\Big(\sum\limits_i m_i\mathbf{z}_i^l*\mathbf{W}_{G_{i,j}}^l\Big),
\end{equation}
\end{small}

\vspace{-1em}
\noindent where $\mathbf{z}_i^l$ and $\mathbf{z}_j^{l+1}$ are the $i$-th input feature map and the $j$-th output feature map at the $l$-the layer, respectively. $\mathbf{W}_{G_{i,j}}^l$ represents the 2D kernel of $i$-th input channel in the $j$-th filter at the $l$-th layer. $*$ and $f(\cdot)$ refer to convolutional operator and non-linearity (ReLU), respectively. 
After training, we remove the feature maps with a zero soft mask that are associated with the corresponding channels at the current layer and the filters at the upper layer. 

\section{Experiments}

\subsection{Experimental Settings}
We evaluate the proposed GAL approach on three widely-used datasets, MNIST \cite{lecun1998gradient}, CIFAR-10 \cite{krizhevsky2009learning} and ImageNet ILSVRC 2012 \cite{russakovsky2015imagenet}. 
We use channel selection to prune plain networks (LeNet \cite{lecun1998gradient} and VGGNet \cite{simonyan2015very}) and DenseNets \cite{huang2017densely}, branch selection for GoogLeNet \cite{szegedy2015going}, and block selection for ResNets \cite{he2016deep}. For ResNets, we also leverage channel selection to block selection that jointly prunes these heterogeneous structures to largely improve the performance of the pruned network.

\textbf{Implementations:} We use PyTorch \cite{paszke2017automatic} to implement GAL. We solve the optimization problem of Eq. (\ref{eq1}) by running on two NVIDIA GTX 1080Ti GPUs with 128GB of RAM. The weight decay is set to 0.0002 and the momentum is set to 0.9. The hyper-parameter $\lambda$ is selected by cross-validation in the range [0.01, 0.1] for channel pruning on LeNet, VGGNet and DenseNets, and the range [0.1, 1] for branch and block pruning on GoogLeNet and ResNets. The drop rate in dropout is set to 0.1. The other training parameters are discussed in different datasets in Section \ref{sec_c}. 

\textbf{Discriminator Architecture:} The discriminator $D$ plays a very important role in striking a balance between simplicity and network capacity to avoid being trivially fooled. In this paper, we select a unified and relative simple architecture, which is composed of three fully-connected (FC) layers and non-linearity (ReLU) with the neurons of 128-256-128. The input is the features from the baseline $f_b(x)$ and the pruned network $f_g(x)$, while the output is the binary prediction to predict the input from baseline or pruned network. 

\subsection{Comparison with the State-of-the-art}
\label{sec_c}

\begin{table}[t]
\scriptsize
\centering
\begin{tabular}{ccccc}
\Xhline{0.1em}
Model & Error/+FT \% & FLOPs(PR) & \#Param.(PR) & \#Filter/Node \\
\Xhline{0.1em}
LeNet & 0.8 & 2.29M(0\%) & 0.43M(0\%) & 20-50-500 \\
SSL \cite{wen2016learning} & -/1.00 & 0.20M(91.3\%) & 0.10M(76.7\%) & 3-12-500 \\
NISP \cite{yu2018nisp} & -/0.82 & 0.65M(71.6\%) & 0.11M(74.4\%) & 10-25-250 \\
GAL-0.01 & 0.95/0.86 & 0.43M(81.2\%) & 0.05M(88.4\%) & 10-15-198 \\
GAL-0.05 & 1.05/0.90 & 0.17M(92.6\%) & 0.03M(93.0\%) & 4-13-121 \\
GAL-0.1 & 1.03/1.01 & 0.10M(95.6\%) & 0.03M(93.0\%) & 2-15-106 \\
\Xhline{0.1em}
\end{tabular}
\vspace{.1em}
\caption{Pruning results of LeNet on MNIST. In all tables and figures, Error/+FT means error without/with fine-tuning, PR represents the pruned rate, GAL-$\lambda$ refers to GAL with sparsity factor $\lambda$ and M/B means million/billion.}
\label{tab_LeNet}
\end{table}

\subsubsection{MNIST}
We evaluate the effectiveness of GAL on MNIST in LeNet. For training parameters, we apply GAL with three groups of hyper-parameter $\lambda$ (0.01, 0.05 and 0.1) with the mini-batch size of 128 for 100 epochs. The initial learning rate is set to 0.001 and is scaled by 0.1 over 40 epochs. As shown in Table \ref{tab_LeNet}, compared to SSL \cite{wen2016learning} and NISP \cite{yu2018nisp}, GAL achieves the best trade-off between FLOPs/parameter pruned rate and the classification error. For example, by setting $\lambda$ to 0.05, the error of GAL only increases by 0.1\% with 92.6\% and 93\% pruned rate in FLOPs and parameter, respectively. In addition, we found that fine-tuning the pruned LeNet with GAL only achieves a limited decrease in error. Fine-tuning instead increases the error when $\lambda$ is set to 0.1. This is due to the fact that the output features learned by GAL have already had a strong discriminability, which may be reduced by fine-tuning. 

\begin{table}[t]
\footnotesize
\centering
\begin{tabular}{cccc}
\Xhline{0.1em}
Model & Error/+FT \% & FLOPs(PR) & \#Param.(PR) \\
\Xhline{0.1em}
VGGNet & 6.04 & 313.73M(0\%) & 14.98M(0\%) \\
L1 \cite{li2017pruning} & -/6.60 & 206.00M(34.3\%) & 5.40M(64.0\%) \\
SSS*\cite{huang2018data} & 6.37\% & 199.93M(36.3\%) & 4.99M(66.7\%) \\
SSS*\cite{huang2018data} & 6.98\% & 183.13M(41.6\%) & 3.93M(73.8\%) \\
GAL-0.05 & 7.97/6.23 & 189.49M(39.6\%) & 3.36M(77.6\%) \\
GAL-0.1 & 9.22/6.58 & 171.89M(45.2\%) & 2.67M(82.2\%) \\
\Xhline{0.1em}
\end{tabular}
\vspace{.3em}
\caption{Pruning results of VGGNet on CIFAR-10. SSS* is the results based on our implementation}
\label{tab_VGGNet}
\end{table}

\subsubsection{CIFAR-10}
We further evaluate the performance of the proposed GAL on CIFAR-10 in five popular networks, VGGNet, DenseNet-40, GoogLeNet, ResNet-56 and ResNet-110. For VGGNet, we take a variation of the original VGG-16 for CIFAR-10 from \cite{li2017pruning,Zagoruyko}. DenseNet-40 has 40 layers with growth rate 12. For GoogLeNet, we also take a variation of the original GoogLeNet by changing the final output class number for CIFAR-10.

\textbf{VGGNet:} The baseline achieves the classification error 6.04\%. GAL is applied to prune it with the mini-batch size of 128 for 100 epochs. The initial learning rate is set to 0.01, and is scaled by 0.1 over 30 epochs. As shown in Table \ref{tab_VGGNet}, compared to L1 \cite{li2017pruning} and SSS \cite{huang2018data}, our GAL achieves a lowest error and highest pruned rate in both FLOPs and parameters. For example, GAL with setting $\lambda$ to 0.05 achieves the lowest error (6.23\% \emph{vs.} 6.60\% by L1 and 6.37\% by SSS) by the highest pruned rate of FLOPs (39.6\% \emph{vs.} 34.3\% by L1 and 36.3\% by SSS) and parameters (77.6\% \emph{vs.} 64.0\% by L1 and 73.8\% by SSS).

\begin{table}[t]
\scriptsize
\centering
\begin{tabular}{cccc}
\Xhline{0.1em}
Model & Error/+FT \% & FLOPs(PR) & \#Param.(PR) \\
\Xhline{0.1em}
DenseNet-40 & 5.19 & 282.92M(0\%) & 1.04M(0\%) \\
Liu \emph{et al.}-40\% \cite{liu2017learning} & -/5.19 & 190M(32.8\%) & 0.66M(36.5\%) \\
Liu \emph{et al.}-70\% \cite{liu2017learning} & -/5.65 & 120M(57.6\%) & 0.35M(66.3\%) \\
GAL-0.01 & 5.71/5.39 & 182.92M(35.3\%) & 0.67M(35.6\%) \\
GAL-0.05 & 6.47/5.50 & 128.11M(54.7\%) & 0.45M(56.7\%) \\
GAL-0.1 & 8.1/6.77 & 80.89M(71.4\%) & 0.26M(75.0\%) \\
\Xhline{0.1em}
\end{tabular}
\vspace{.3em}
\caption{Pruning results of DenseNet-40 on CIFAR-10. Liu \emph{et al.}-$\alpha$\% means about $\alpha$ percentage of parameters are pruned.}
\label{tab_densenet}
\end{table}

\textbf{DenseNet-40:}
According to the principle of channel selection in Section \ref{SS}, we should prune the input channels at the current layer and the corresponding output feature maps and the filters at the upper layer in DenseNets. But this leads to a mismatch of the dimension in the following layers. This is due to the complex dense connectivity of each layer in DenseNets. We therefore only prune the input channels in DenseNet-40, as suggested in  \cite{liu2017learning}. The training setup is the same to VGGNet, except the mini-batch size is 64. 
The pruning results of DenseNet-40 are summarized in Table \ref{tab_densenet}. GAL achieves a comparable result with Liu \emph{et al.} \cite{liu2017learning}. For example, when $\lambda$ is set to 0.01, 3362 out of 8904 channels are pruned by GAL with a higher computational saving of (35.3\% \emph{vs.} 32.8\%), but only with a slightly higher error (5.39\% \emph{vs.} 5.19\%), compared to Liu \emph{et al.}-40\%. 

\begin{table}[t]
\footnotesize
\centering
\begin{tabular}{cccc}
\Xhline{0.1em}
Model & Error/+FT \% & FLOPs(PR) & \#Param.(PR) \\
\Xhline{0.1em}
GoogLeNet & 4.95 & 1.52B(0\%) & 6.15M(0\%) \\
Random & -/5.46 & 0.96B(36.8\%) & 3.58M(41.8\%) \\
L1* \cite{li2017pruning} & -/5.46 & 1.02B(32.9\%) & 3.51M(42.9\%) \\
APoZ* \cite{hu2016network} &  -/7.89 & 0.76B(50.0\%) & 2.85M(53.7\%) \\
GAL-0.5 & 6.07/5.44 & 0.94B(38.2\%) & 3.12M(49.3\%) \\
\Xhline{0.1em}
\end{tabular}
\vspace{.3em}
\caption{Pruning results of GoogLeNet on CIFAR-10. L1* and APoZ* are the results based on our implementation.}
\label{tab_googlenet}
\end{table}

\textbf{GoogLeNet:}
For better comparison, we re-implemented L1 \cite{li2017pruning} and APoZ \cite{hu2016network} on GoogLeNet and also introduce random pruning, because of lack of pruning results on GoogLeNet in CIFAR-10. For Random, L1 and APoZ, we simply prune the same number of branches in each inception module based on their pruning criteria as GAL-0.5 for a fair comparison. The training parameters of GAL are the same to prune DenseNet-40 (not including $\lambda$) and the first convolutional layer is skipped to add the soft mask. As presented in Table \ref{tab_googlenet}, GAL achieves the best trade-off by removing 14 of 36 branches with a rate of FLOPs saving of 38.2\%, parameter saving of 49.3\% and only an increase of 0.49\% classification error, compared to all methods. 
This is because GAL employs the more flexible branch selection by learning the soft mask than L1 and APoZ based on the statistical property. 
Note that the simplest random approach works reasonably well, which is possibly due to the self-recovery ability of the distributed representations. 
In addition, the branches of $3\times3$ convolutional filters with a large number of parameters are more removed by APoZ, which leads to significant FLOPs and parameters reduction and also significant error increase.

\begin{table}[t]
\footnotesize
\centering
\begin{tabular}{cccc}
\Xhline{0.1em}
Model & Error/+FT \% & FLOPs(PR) & \#Param.(PR) \\
\Xhline{0.1em}
ResNet-56 & 6.74 & 125.49M(0\%) & 0.85M(0\%) \\
He \emph{et al.} \cite{he2017channel} & 9.20/8.20 & 62M(50.6\%) & - \\
L1 \cite{li2017pruning} & -/6.94 & 90.9M(27.6\%) & 0.73M(14.1\%) \\
NISP \cite{yu2018nisp} & -/6.99 & 81M(35.5\%) & 0.49M(42.4\%) \\
GAL-0.6 & 7.02/6.62 & 78.30M(37.6\%) & 0.75M(11.8\%) \\
GAL-0.8 & 9.64/8.42 & 49.99M(60.2\%) & 0.29M(65.9\%) \\
\Xhline{0.1em}
ResNet-110 & 6.5 & 252.89M(0\%) & 1.72(0\%) \\
\multirow{2}*{L1 \cite{li2017pruning}} & -/6.45 & 213M(15.8\%) & 1.68M(2.3\%) \\
& -/6.7 & 155M(38.7\%) & 1.16M(32.6\%) \\
GAL-0.1 & 7.45/6.41 & 205.7M(18.7\%) & 1.65M(4.1\%) \\
GAL-0.5 & 7.45/7.26 & 130.2M(48.5\%) & 0.95M(44.8\%) \\
\Xhline{0.1em}
\end{tabular}
\vspace{.3em}
\caption{Pruning results of ResNet-56/110 on CIFAR-10.}
\label{tab_resnet_56_110}
\end{table}

\textbf{ResNets:} 
To evaluate the effectiveness of block selection in GAL, we use ResNet-56 and ResNet-110 as our baseline models. The training parameters of GAL on both ResNet-56 and ResNet-110 are the same to prune VGGNet (not including $\lambda$) and the first convolutional layer is also skipped to add the soft mask. 
The pruning results of both ResNet-56 and ResNet-110 are summarized in Table \ref{tab_resnet_56_110}. 
For ResNet-56, when $\lambda$ is set to 0.6, 10 out of 27 residual blocks are removed by GAL, which achieves a 37.6\% pruned rate in FLOPs while with a decrease of 0.12\% error. 
This indicates that there are redundant residual blocks in ResNet-56. 
Moreover, compared to L1 \cite{li2017pruning} and NISP \cite{yu2018nisp}, GAL-0.6 also achieves the best performance. 
When more residual blocks are pruned (16 when $\lambda$ is set to 0.8), GAL-0.8 still achieves the higher pruned rate in FLOPs (60.2\% \emph{vs.} 50.6\%), with a slightly higher classification error (8.42\% \emph{vs.} 8.20\%) compared to He \emph{et al.} \cite{he2017channel}.
For ResNet-110, compared to L1, GAL achieves better results by pruning 10 out of 54 residual blocks, when $\lambda$ is set to 0.1. 
After optimization for ResNet-56 and ResNet-110, the bottom residual blocks are easier to prune. To explain, top blocks often have high-level semantic information that is necessary for maintaining the classification accuracy.

\begin{table}[t]
\small
\centering
\begin{tabular}{ccccc}
\Xhline{0.1em}
Model & Top-1 \% & Top-5 \% & FLOPs & \#Param. \\
\Xhline{0.1em}
ResNet-50 & 23.85 & 7.13 & 4.09B & 25.5M \\
ThiNet-50 \cite{luo2017ThiNet} & 28.99 & 9.98 & 1.71B & 12.38M \\
ThiNet-30 \cite{luo2017ThiNet} & 31.58 & 11.70 & 1.10B & 8.66M \\
He \emph{et al.} \cite{he2017channel} & 27.70 & 9.20 & 2.73B & - \\
GDP-0.6 \cite{lin2018accelerating} & 28.81 & 9.29 & 1.88B & - \\
GDP-0.5 \cite{lin2018accelerating} & 30.42 & 9.86 & 1.57B & - \\
SSS-32 \cite{huang2018data} & 25.82 & 8.09 & 2.82B & 18.6M \\
SSS-26 \cite{huang2018data} & 28.18 & 9.21 & 2.33B & 15.6M \\
GAL-0.5 & 28.05 & 9.06 & 2.33B & 21.2M\\
GAL-1 & 30.12& 10.25 & 1.58B & 14.67M\\
GAL-0.5-joint& 28.20 & 9.18 & 1.84B & 19.31M \\
GAL-1-joint & 30.69 & 10.88 & 1.11B & 10.21M \\
\Xhline{0.1em}
\end{tabular}
\vspace{.3em}
\caption{Pruning results of ResNet-50 on ImageNet. X-joint means jointly pruning heterogeneous structures (channels and blocks).}
\label{tab_inception_resnet}
\end{table}

\subsubsection{ImageNet ILSVRC 2012}
GAL was also evaluated on ImageNet using ResNet-50. 
We train the pruned network with the mini-batch size of 32 for 30 epochs. The initial learning rate is set to 0.01 and is scaled by 0.1 over 10 epochs. As shown in Table \ref{tab_inception_resnet}, GAL without jointly pruning blocks and channels is able to obtain $1.76\times$ and $2.59\times$ speedup (FLOPs rate) (2.33B and 1.58B \emph{vs.} 4.09B in ResNet-50) by setting $\lambda$ to 0.5 and 1, with an increase of 1.93\% and 3.12\% in Top-5 error, respectively. 
However, GAL-0.5 and GAL-1 only achieve a $1.2\times$ and $1.74\times$ parameter compression rate, which is due to the fact that most of the pruned blocks comes from the bottom layers with a small number of parameters. 
By jointly pruning blocks and channels, we achieve a higher speedup and compression. 
For example, compared to GAL-0.5, GAL-0.5-joint achieves the higher speedup and compression by a factor of $2.22\times$ and $1.32\times$ (\emph{vs.} $1.75\times$ and $1.2\times$), respectively.  Furthermore, compared to SSS-26 \cite{huang2018data}, He \emph{et al} \cite{he2017channel} and GDP-0.6 \cite{lin2018accelerating}, GAL-0.5-joint also achieves the best trade-off between Top-5 error and speedup. 
With almost the same speedup, our GAL-1-joint outperforms ThiNet-30 \cite{luo2017ThiNet} by 0.89\% and 0.82\% in Top-1 and Top-5 error, respectively.

\subsection{Ablation Study}
To evaluate the effectiveness of GAL, which lies in adversarial regularization, FISTA and GANs, we select ResNet-56 and DenseNet-40 for an ablation study. 

\subsubsection{Effect of the Regularizers on Discriminator $D$}
\begin{figure}[!t]
\centering
  \subfigure{
    \includegraphics[scale = 0.35]{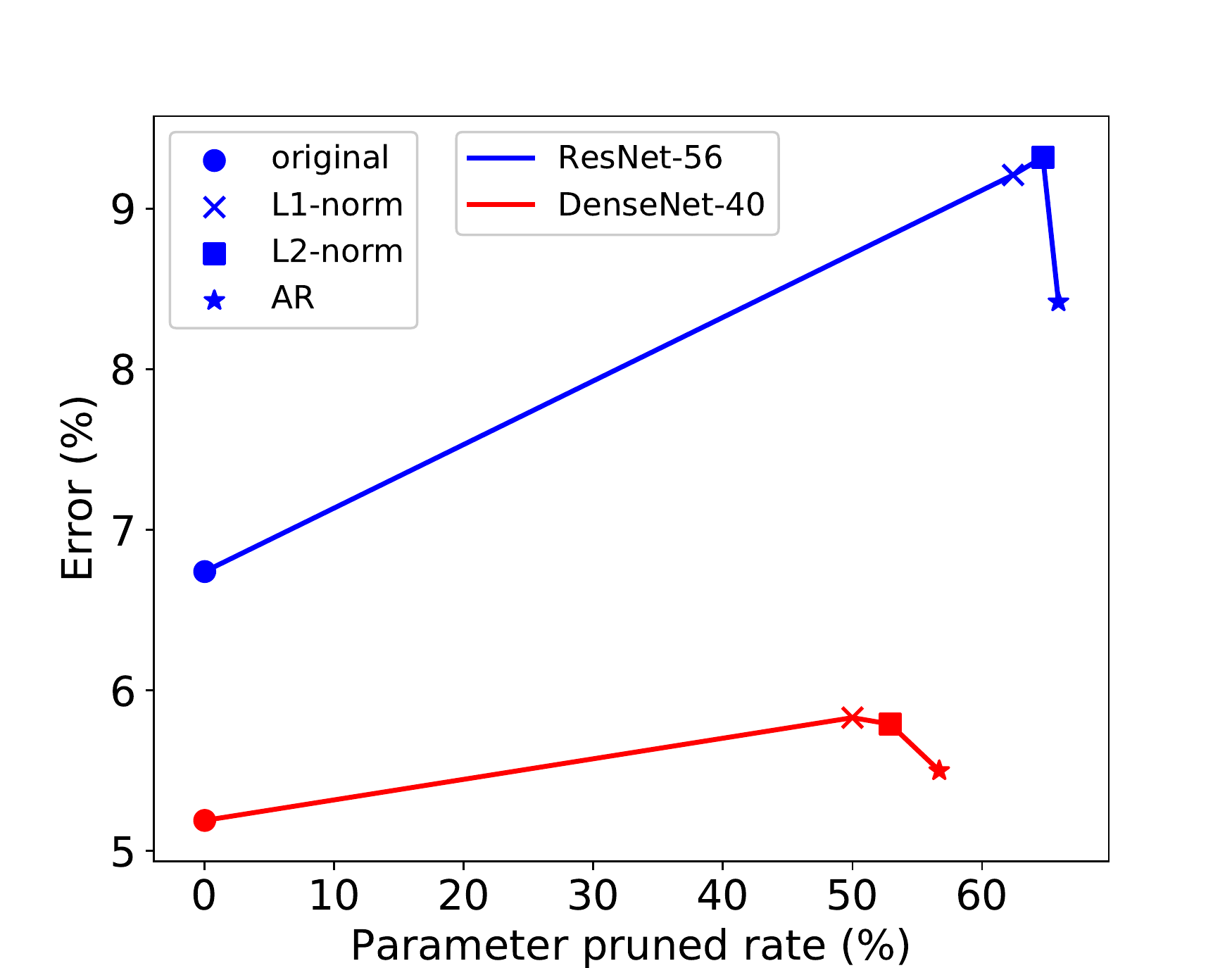}
    \label{fig_param}
  }
 \vspace{.3em}
 \caption{Comparison of the different discriminator regularizers on ResNet-56 and DenseNet-40.} 
 \label{fig_regularization}
\end{figure}

We train our GAL approach with three types of discriminator regularizers, L1-norm, L2-norm and adversarial regularization (AR). For a fair comparison, all the training parameters are the same. As shown in Fig. \ref{fig_regularization}, adversarial regularization achieves the best performance, compared to the L1-norm and L2-norm. This is because AR prolongs the competition between generator and discriminator to achieve the better output features of generator, which are close to baseline and fool the discriminator.

\begin{table}[t]
\centering
\scriptsize
\begin{tabular}{cccc}
\Xhline{0.1em}
Model & Error/PN/PN-FT\% & FLOPs(PR) & \#Param.(PR) \\
\Xhline{0.1em}
ResNet-56 & 6.74 & 125.49M(0\%) & 0.85M(0\%) \\
GAL-AR-SGD & 9.67/89.14/9.65 & 50.27M(59.9\%) & 0.59M(30.6\%) \\
Random & -/89.96/12.32 & 50.27M(59.9\%) & 0.59M(30.6\%) \\
GAL-AR-FISTA & 9.64/9.64/8.42 & 49.99M(60.2\%) & 0.29M(65.9\%) \\
\Xhline{0.1em}
DenseNet-40 & 5.19 & 282.92M(0\%) & 1.04M(0\%) \\
GAL-AR-SGD & 6.76/64.58/7.64 & 140.55M(50.3\%) & 0.46M(55.8\%) \\
Random & -/89.23/11.08 & 140.55M(50.3\%) & 0.46M(55.8\%) \\
GAL-AR-FISTA & 6.47/6.47/5.50 & 128.11M(54.7\%) & 0.45M(56.7\%) \\
\Xhline{0.1em}
\end{tabular}
\caption{Results of the different optimizers. PN/PN-FT is the pruned networks without/with fine-tuning. Random means training the architecture (same to SGD) from scratch.}
\label{tab_ablation_optimizer}
\end{table}

\subsubsection{Effect on the Optimizers}
We compare our FISTA with SGD optimizer. For SGD, we cannot obtain the soft mask with an exact scaling factor of 0. Therefore, a hard threshold is required in the pruning stage. We set the threshold to 0.0001 in our experiments. 
As presented in Table \ref{tab_ablation_optimizer}, 
compared to the random method, SGD achieves a lower error with the same architecture. It indicates that SGD provides better initial values for the pruned network (PN). After pruning with thresholding, the accuracy drops significantly (See the columns of Error and PN in Table \ref{tab_ablation_optimizer}), as the pruned small near-zero weights might have large impact on the final network output. Advantageously, GAL with FISTA can safely remove the redundant structures in the training process, and achieves better performance compared to SGD. 

\subsubsection{Effect of the GANs}
\begin{figure}[!t]
\centering
\hspace{-1em}
  \subfigure[ResNet-56]{
    \includegraphics[scale = 0.282]{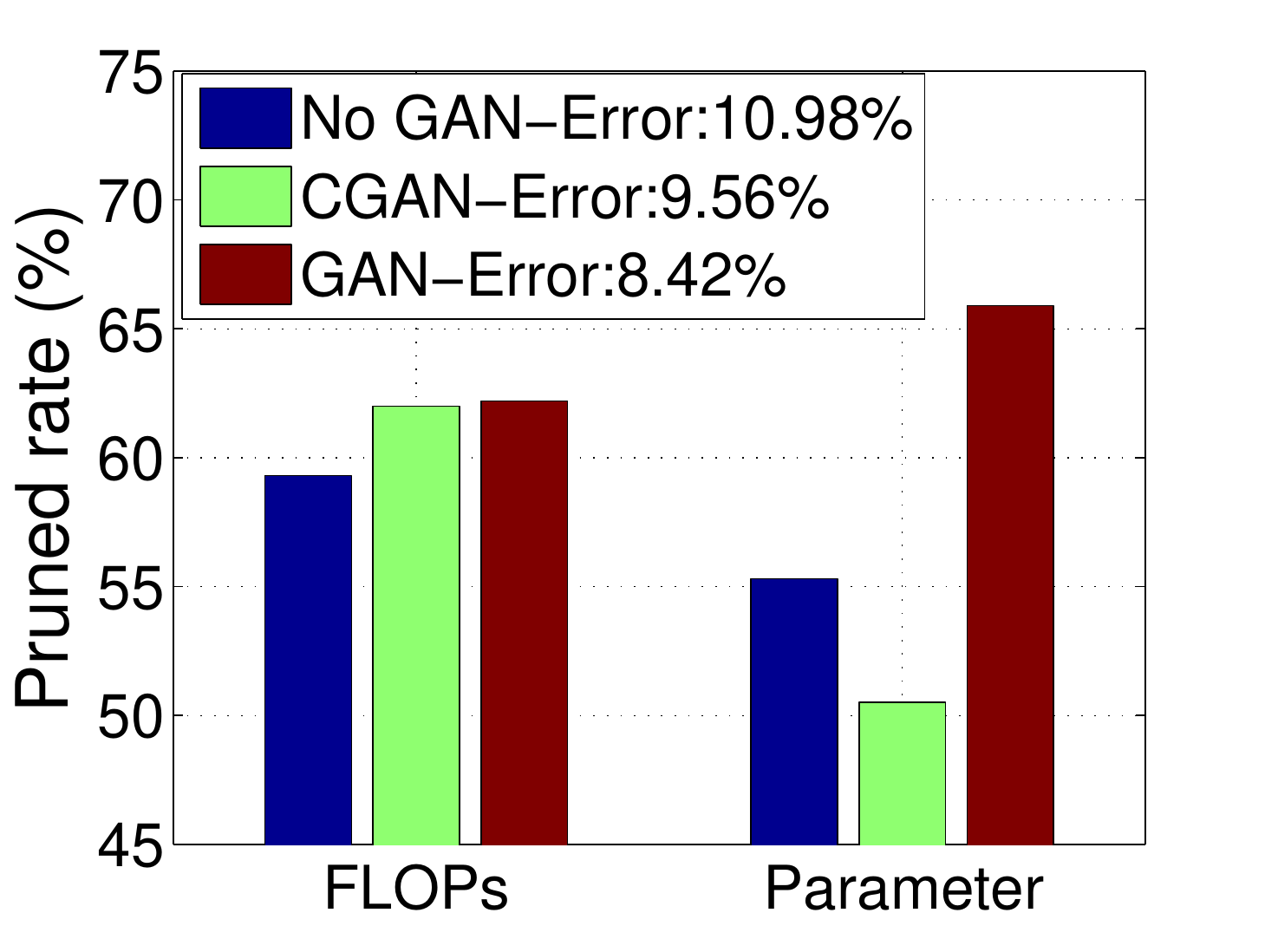}
    \label{fig_flops}
    }
    \hspace{-1em}
  \subfigure[DenseNet-40]{
    \includegraphics[scale = 0.282]{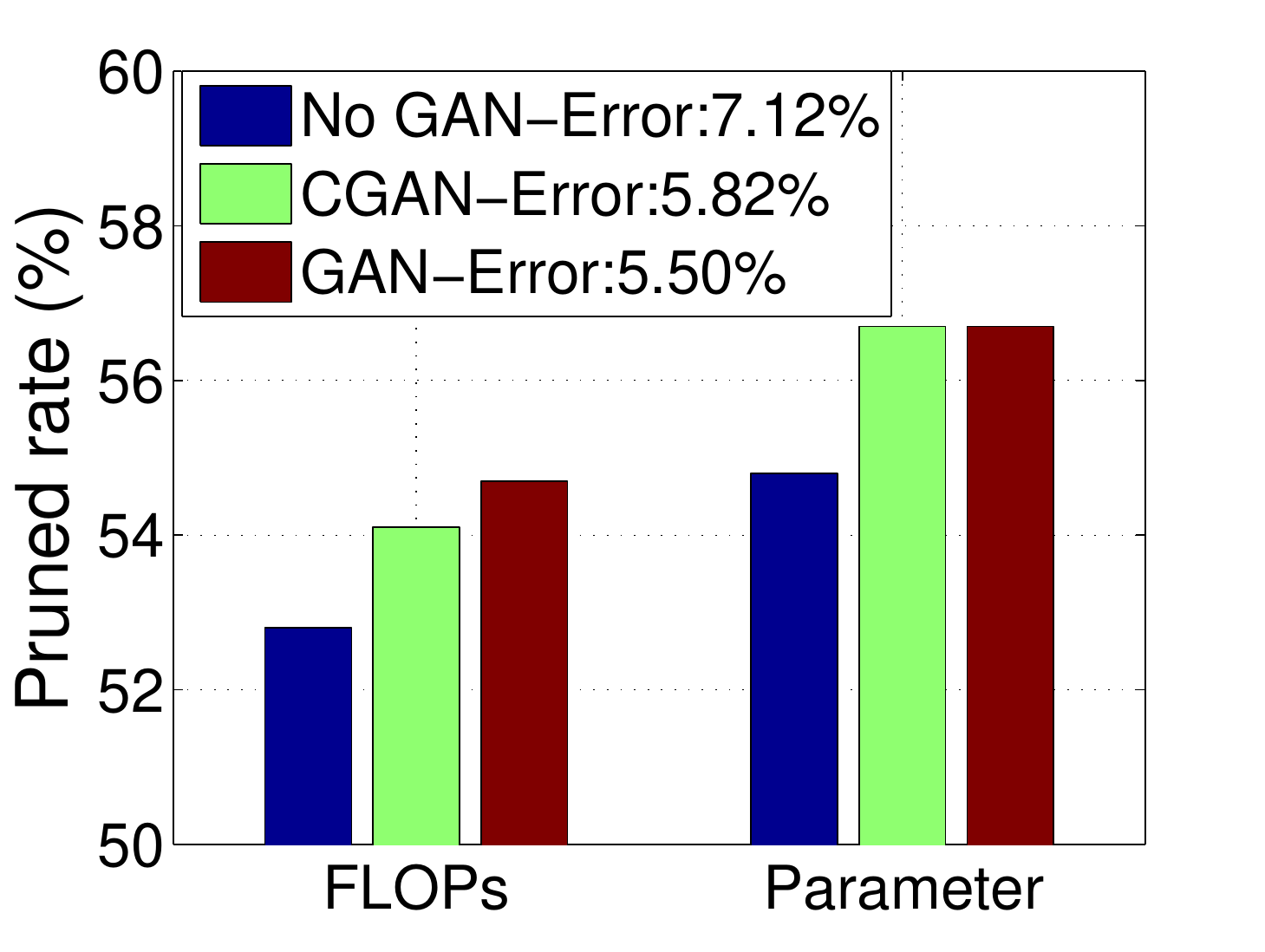}
    \label{fig_param}
  }
 \caption{Comparison of GANs on ResNet-56 and DenseNet-40.} 
 \label{fig_gan}
\end{figure}

We train the pruned network with and without the GAN, and also make a comparison with CGAN \cite{mirza2014conditional} by using the FISTA. 
For training CGAN, we only need to modify the adversarial loss function in Eq. (\ref{eq2}) by the loss of CGAN, and the optimization with related training parameters are same as GAL. 
The results are summarized in Fig. \ref{fig_gan}.
First, the lack of GANs leads to significant error increase. Second, the GAN achieves a better result than CGAN.
For example, with the same regularization and optimizer on ResNet-56, label-free GAL achieves a 8.42\% error with a 65.9\% parameter pruned rate \emph{vs.} 9.56\% error with 50.5\% parameter pruned rate in label-dependent CGAN. We conjecture this is due to the class label that is added to the discriminator in CGAN, which instead affects the output features of generator to approximate baseline during training.

\section{Conclusion}
In this paper, we developed a generative adversarial learning (GAL) approach to effectively structured prune CNNs, which jointly prunes heterogeneous structures in an end-to-end manner. 
We introduced a soft mask to scale the output of specific structures, upon which a new objective function with $\ell_1$-regularization on the soft mask is designed to align the output of the baseline and the network with this mask. 
To effectively solve the optimization problem, we used a label-free generative adversarial learning to learn the pruned network with the sparse soft mask. 
Moreover, by forcing more scaling factors in the soft mask to zero, we leverage the fast iterative shrinkage-thresholding algorithm to quickly and reliably remove the corresponding redundant structures. 
We have comprehensively evaluated the performance of the proposed approach on a variety of state-of-the-art CNN architectures over different datasets, which demonstrates the superior performance gains over the state-of-the-art methods.


{\small
\bibliographystyle{ieee_fullname}
\bibliography{egbib}

\begin{thebibliography}{10}\itemsep=-1pt

\bibitem{baker2017designing}
Bowen Baker, Otkrist Gupta, Nikhil Naik, and Ramesh Raskar.
\newblock Designing neural network architectures using reinforcement learning.
\newblock In {\em ICLR}, 2017.

\bibitem{beck2009fast}
Amir Beck and Marc Teboulle.
\newblock A fast iterative shrinkage-thresholding algorithm for linear inverse
  problems.
\newblock {\em SIAM journal on imaging sciences}, 2(1):183--202, 2009.

\bibitem{denton2014exploiting}
Emily~L Denton, Wojciech Zaremba, Joan Bruna, Yann LeCun, and Rob Fergus.
\newblock Exploiting linear structure within convolutional networks for
  efficient evaluation.
\newblock In {\em NIPS}, pages 1269--1277, 2014.

\bibitem{girshick2015fast}
Ross Girshick.
\newblock Fast r-cnn.
\newblock In {\em ICCV}, pages 1440--1448, 2015.

\bibitem{girshick2014rich}
Ross Girshick, Jeff Donahue, Trevor Darrell, and Jitendra Malik.
\newblock Rich feature hierarchies for accurate object detection and semantic
  segmentation.
\newblock In {\em CVPR}, pages 580--587, 2014.

\bibitem{goldstein2014field}
Tom Goldstein, Christoph Studer, and Richard Baraniuk.
\newblock A field guide to forward-backward splitting with a fasta
  implementation.
\newblock {\em arXiv preprint arXiv:1411.3406}, 2014.

\bibitem{goodfellow2014generative}
Ian Goodfellow, Jean Pouget-Abadie, Mehdi Mirza, Bing Xu, David Warde-Farley,
  Sherjil Ozair, Aaron Courville, and Yoshua Bengio.
\newblock Generative adversarial nets.
\newblock In {\em NIPS}, pages 2672--2680, 2014.

\bibitem{guo2016dynamic}
Yiwen Guo, Anbang Yao, and Yurong Chen.
\newblock Dynamic network surgery for efficient dnns.
\newblock In {\em NIPS}, pages 1379--1387, 2016.

\bibitem{han2016eie}
Song Han, Xingyu Liu, Huizi Mao, Jing Pu, Ardavan Pedram, Mark~A Horowitz, and
  William~J Dally.
\newblock Eie: efficient inference engine on compressed deep neural network.
\newblock In {\em ISCA}, pages 243--254, 2016.

\bibitem{han2016deep}
Song Han, Huizi Mao, and William~J Dally.
\newblock Deep compression: Compressing deep neural network with pruning,
  trained quantization and huffman coding.
\newblock In {\em ICLR}, 2016.

\bibitem{han2015learning}
Song Han, Jeff Pool, John Tran, and William Dally.
\newblock Learning both weights and connections for efficient neural network.
\newblock In {\em NIPS}, pages 1135--1143, 2015.

\bibitem{hassibi1993second}
Babak Hassibi and David~G Stork.
\newblock Second order derivatives for network pruning: Optimal brain surgeon.
\newblock In {\em NIPS}, 1993.

\bibitem{he2016deep}
Kaiming He, Xiangyu Zhang, Shaoqing Ren, and Jian Sun.
\newblock Deep residual learning for image recognition.
\newblock In {\em CVPR}, pages 770--778, 2016.

\bibitem{he2018amc}
Yihui He, Ji Lin, Zhijian Liu, Hanrui Wang, Li-Jia Li, and Song Han.
\newblock Amc: Automl for model compression and acceleration on mobile devices.
\newblock In {\em ECCV}, pages 784--800, 2018.

\bibitem{he2018pruning}
Yang He, Ping Liu, Ziwei Wang, and Yi Yang.
\newblock Pruning filter via geometric median for deep convolutional neural
  networks acceleration.
\newblock {\em arXiv preprint arXiv:1811.00250}, 2018.

\bibitem{he2017channel}
Yihui He, Xiangyu Zhang, and Jian Sun.
\newblock Channel pruning for accelerating very deep neural networks.
\newblock In {\em ICCV}, 2017.

\bibitem{hinton2015distilling}
Geoffrey Hinton, Oriol Vinyals, and Jeff Dean.
\newblock Distilling the knowledge in a neural network.
\newblock {\em arXiv preprint arXiv:1503.02531}, 2015.

\bibitem{howard2017mobilenets}
Andrew~G Howard, Menglong Zhu, Bo Chen, Dmitry Kalenichenko, Weijun Wang,
  Tobias Weyand, Marco Andreetto, and Hartwig Adam.
\newblock Mobilenets: Efficient convolutional neural networks for mobile vision
  applications.
\newblock {\em arXiv preprint arXiv:1704.04861}, 2017.

\bibitem{hu2016network}
Hengyuan Hu, Rui Peng, Yu-Wing Tai, and Chi-Keung Tang.
\newblock Network trimming: A data-driven neuron pruning approach towards
  efficient deep architectures.
\newblock {\em arXiv preprint arXiv:1607.03250}, 2016.

\bibitem{huang2017densely}
Gao Huang, Zhuang Liu, Laurens van~der Maaten, and Kilian~Q Weinberger.
\newblock Densely connected convolutional networks.
\newblock In {\em CVPR}, pages 4700--4708, 2017.

\bibitem{huang2018data}
Zehao Huang and Naiyan Wang.
\newblock Data-driven sparse structure selection for deep neural networks.
\newblock In {\em ECCV}, 2018.

\bibitem{isola2017image}
Phillip Isola, Jun-Yan Zhu, Tinghui Zhou, and Alexei~A Efros.
\newblock Image-to-image translation with conditional adversarial networks.
\newblock In {\em CVPR}, pages 1125--1134, 2017.

\bibitem{jacob2018quantization}
Benoit Jacob, Skirmantas Kligys, Bo Chen, Menglong Zhu, Matthew Tang, Andrew
  Howard, Hartwig Adam, and Dmitry Kalenichenko.
\newblock Quantization and training of neural networks for efficient
  integer-arithmetic-only inference.
\newblock In {\em CVPR}, pages 2704--2713, 2018.

\bibitem{kim2016compression}
Yong-Deok Kim, Eunhyeok Park, Sungjoo Yoo, Taelim Choi, Lu Yang, and Dongjun
  Shin.
\newblock Compression of deep convolutional neural networks for fast and low
  power mobile applications.
\newblock In {\em ICLR}, 2016.

\bibitem{krizhevsky2009learning}
Alex Krizhevsky and Geoffrey Hinton.
\newblock Learning multiple layers of features from tiny images.
\newblock Technical report, Citeseer, 2009.

\bibitem{krizhevsky2012imagenet}
Alex Krizhevsky, Ilya Sutskever, and Geoffrey~E Hinton.
\newblock Imagenet classification with deep convolutional neural networks.
\newblock In {\em NIPS}, pages 1097--1105, 2012.

\bibitem{lebedev2015speeding}
Vadim Lebedev, Yaroslav Ganin, Maksim Rakhuba, Ivan Oseledets, and Victor
  Lempitsky.
\newblock Speeding-up convolutional neural networks using fine-tuned
  cp-decomposition.
\newblock In {\em ICLR}, 2015.

\bibitem{lebedev2016fast}
Vadim Lebedev and Victor Lempitsky.
\newblock Fast convnets using group-wise brain damage.
\newblock In {\em CVPR}, pages 2554--2564, 2016.

\bibitem{lecun1998gradient}
Yann LeCun, L{\'e}on Bottou, Yoshua Bengio, and Patrick Haffner.
\newblock Gradient-based learning applied to document recognition.
\newblock {\em Proceedings of the IEEE}, 86(11):2278--2324, 1998.

\bibitem{lecun1989optimal}
Yann LeCun, John~S Denker, Sara~A Solla, Richard~E Howard, and Lawrence~D
  Jackel.
\newblock Optimal brain damage.
\newblock In {\em NIPS}, volume~2, pages 598--605, 1989.

\bibitem{li2017pruning}
Hao Li, Asim Kadav, Igor Durdanovic, Hanan Samet, and Hans~Peter Graf.
\newblock Pruning filters for efficient convnets.
\newblock In {\em ICLR}, 2017.

\bibitem{lin2018holistic}
Shaohui Lin, Rongrong Ji, Chao Chen, Dacheng Tao, and Jiebo Luo.
\newblock Holistic cnn compression via low-rank decomposition with knowledge
  transfer.
\newblock {\em PAMI}, 2018.

\bibitem{lin2016towards}
Shaohui Lin, Rongrong Ji, Xiaowei Guo, and Xuelong Li.
\newblock Towards convolutional neural networks compression via global error
  reconstruction.
\newblock In {\em IJCAI}, 2016.

\bibitem{lin2019towards}
Shaohui Lin, Rongrong Ji, Yuchao Li, Cheng Deng, and Xuelong Li.
\newblock Towards compact convnets via structure-sparsity regularized filter
  pruning.
\newblock {\em arXiv preprint arXiv:1901.07827}, 2019.

\bibitem{lin2018accelerating}
Shaohui Lin, Rongrong Ji, Yuchao Li, Yongjian Wu, Feiyue Huang, and Baochang
  Zhang.
\newblock Accelerating convolutional networks via global \& dynamic filter
  pruning.
\newblock In {\em IJCAI}, 2018.

\bibitem{liu2017learning}
Zhuang Liu, Jianguo Li, Zhiqiang Shen, Gao Huang, Shoumeng Yan, and Changshui
  Zhang.
\newblock Learning efficient convolutional networks through network slimming.
\newblock In {\em ICCV}, pages 2755--2763, 2017.

\bibitem{luo2017ThiNet}
Jianhao Luo, Jianxin Wu, and Weiyao Lin.
\newblock Thinet: A filter level pruning method for deep neural network
  compression.
\newblock In {\em ICCV}, 2017.

\bibitem{mirza2014conditional}
Mehdi Mirza and Simon Osindero.
\newblock Conditional generative adversarial nets.
\newblock {\em arXiv preprint arXiv:1411.1784}, 2014.

\bibitem{molchanov2017pruning}
Pavlo Molchanov, Stephen Tyree, Tero Karras, Timo Aila, and Jan Kautz.
\newblock Pruning convolutional neural networks for resource efficient
  inference.
\newblock In {\em ICLR}, 2017.

\bibitem{park2017faster}
Jongsoo Park, Sheng Li, Wei Wen, Ping Tak~Peter Tang, Hai Li, Yiran Chen, and
  Pradeep Dubey.
\newblock Faster cnns with direct sparse convolutions and guided pruning.
\newblock In {\em ICLR}, 2017.

\bibitem{paszke2017automatic}
Adam Paszke, Sam Gross, Soumith Chintala, Gregory Chanan, Edward Yang, Zachary
  DeVito, Zeming Lin, Alban Desmaison, Luca Antiga, and Adam Lerer.
\newblock Automatic differentiation in pytorch.
\newblock In {\em NIPS Workshops}, 2017.

\bibitem{rastegari2016xnor}
M. Rastegari, V. Ordonez, J. Redmon, and A. Farhadi.
\newblock Xnor-net: Imagenet classification using binary convolutional neural
  networks.
\newblock In {\em ECCV}, 2016.

\bibitem{real2017large}
Esteban Real, Sherry Moore, Andrew Selle, Saurabh Saxena, Yutaka~Leon Suematsu,
  Jie Tan, Quoc Le, and Alex Kurakin.
\newblock Large-scale evolution of image classifiers.
\newblock In {\em ICML}, 2017.

\bibitem{ren2015faster}
Shaoqing Ren, Kaiming He, Ross Girshick, and Jian Sun.
\newblock Faster r-cnn: Towards real-time object detection with region proposal
  networks.
\newblock In {\em NIPS}, pages 91--99, 2015.

\bibitem{romero2015fitnets}
Adriana Romero, Nicolas Ballas, Samira~Ebrahimi Kahou, Antoine Chassang, Carlo
  Gatta, and Yoshua Bengio.
\newblock Fitnets: Hints for thin deep nets.
\newblock In {\em ICLR}, 2015.

\bibitem{russakovsky2015imagenet}
Olga Russakovsky, Jia Deng, Hao Su, Jonathan Krause, Sanjeev Satheesh, Sean Ma,
  Zhiheng Huang, Andrej Karpathy, Aditya Khosla, Michael Bernstein, et~al.
\newblock Imagenet large scale visual recognition challenge.
\newblock {\em IJCV}, 115(3):211--252, 2015.

\bibitem{simonyan2015very}
Karen Simonyan and Andrew Zisserman.
\newblock Very deep convolutional networks for large-scale image recognition.
\newblock In {\em ICLR}, 2015.

\bibitem{szegedy2015going}
Christian Szegedy, Wei Liu, Yangqing Jia, Pierre Sermanet, Scott Reed, Dragomir
  Anguelov, Dumitru Erhan, Vincent Vanhoucke, and Andrew Rabinovich.
\newblock Going deeper with convolutions.
\newblock In {\em CVPR}, pages 1--9, 2015.

\bibitem{wen2016learning}
Wei Wen, Chunpeng Wu, Yandan Wang, Yiran Chen, and Hai Li.
\newblock Learning structured sparsity in deep neural networks.
\newblock In {\em NIPS}, 2016.

\bibitem{xie2017genetic}
Lingxi Xie and Alan~L Yuille.
\newblock Genetic cnn.
\newblock In {\em ICCV}, pages 1388--1397, 2017.

\bibitem{xie2017aggregated}
Saining Xie, Ross Girshick, Piotr Doll{\'a}r, Zhuowen Tu, and Kaiming He.
\newblock Aggregated residual transformations for deep neural networks.
\newblock In {\em CVPR}, pages 5987--5995, 2017.

\bibitem{ye2018rethinking}
Jianbo Ye, Xin Lu, Zhe Lin, and James~Z Wang.
\newblock Rethinking the smaller-norm-less-informative assumption in channel
  pruning of convolution layers.
\newblock In {\em ICLR}, 2018.

\bibitem{yu2018nisp}
Ruichi Yu, Ang Li, Chun-Fu Chen, Jui-Hsin Lai, Vlad~I Morariu, Xintong Han,
  Mingfei Gao, Ching-Yung Lin, and Larry~S Davis.
\newblock Nisp: Pruning networks using neuron importance score propagation.
\newblock In {\em CVPR}, pages 9194--9203, 2018.

\bibitem{Zagoruyko}
Sergey Zagoruyko.
\newblock 92.45\% on cifar-10 in torch.
\newblock {\em http://torch.ch/blog/2015/07/30/cifar.html}, 2015.

\bibitem{zagoruyko2017paying}
Sergey Zagoruyko and Nikos Komodakis.
\newblock Paying more attention to attention: Improving the performance of
  convolutional neural networks via attention transfer.
\newblock In {\em ICLR}, 2017.

\bibitem{zhang2017interleaved}
Ting Zhang, Guo-Jun Qi, Bin Xiao, and Jingdong Wang.
\newblock Interleaved group convolutions.
\newblock In {\em CVPR}, 2017.

\bibitem{zhang2018shufflenet}
Xiangyu Zhang, Xinyu Zhou, Mengxiao Lin, and Jian Sun.
\newblock Shufflenet: An extremely efficient convolutional neural network for
  mobile devices.
\newblock In {\em CVPR}, 2018.

\bibitem{zhang2015efficient}
Xiangyu Zhang, Jianhua Zou, Xiang Ming, Kaiming He, and Jian Sun.
\newblock Efficient and accurate approximations of nonlinear convolutional
  networks.
\newblock In {\em CVPR}, 2015.

\bibitem{zhuang2018towards}
Bohan Zhuang, Chunhua Shen, Mingkui Tan, Lingqiao Liu, and Ian Reid.
\newblock Towards effective low-bitwidth convolutional neural networks.
\newblock In {\em CVPR}, pages 7920--7928, 2018.

\bibitem{zoph2017neural}
Barret Zoph and Quoc~V Le.
\newblock Neural architecture search with reinforcement learning.
\newblock In {\em ICLR}, 2017.

\bibitem{zoph2018learning}
Barret Zoph, Vijay Vasudevan, Jonathon Shlens, and Quoc~V Le.
\newblock Learning transferable architectures for scalable image recognition.
\newblock In {\em CVPR}, 2018.

\end{thebibliography}
}
\end{document}